\documentclass[sigconf, nonacm]{acmart}

\newcommand\vldbdoi{XX.XX/XXX.XX}
\newcommand\vldbpages{XXX-XXX}
\newcommand\vldbvolume{17}
\newcommand\vldbissue{4}
\newcommand\vldbyear{2023}
\newcommand\vldbauthors{\authors}
\newcommand\vldbtitle{\shorttitle} 
\newcommand\vldbavailabilityurl{https://github.com/knagrecha/saturn}
\newcommand\vldbpagestyle{empty}

\usepackage{graphicx,subfigure,xspace,verbatim,comment}
\usepackage{hyperref,array,color,balance,multirow}
\usepackage{balance,float,url,amsfonts,alltt}
\usepackage{array}
\newcolumntype{P}[1]{>{\centering\arraybackslash}p{#1}}
\usepackage{mathtools,rotating,amsmath}
\usepackage{color,ifpdf,fancyvrb}
\usepackage{etoolbox,listings}
\usepackage{bigstrut,morefloats,pbox}
\usepackage{amsmath}
 \usepackage{balance}
\usepackage{bbm,dsfont}
\usepackage{bbold}
\usepackage{algorithm}
\usepackage{algorithmic}
\usepackage{booktabs}
\usepackage{lipsum}
\usepackage{enumitem}
\usepackage[all]{nowidow}
\usepackage{pifont}
\usepackage{tabularx}
\usepackage{enumitem}
\setlist{
  listparindent=\parindent,
  parsep=0pt,
}

\newcommand{\eat}[1]{}
\newcommand{\system}{\textsc{Saturn}}
\newcommand{\spase}{\textsc{SPASE}}
\newcommand{\red}[1]{\textcolor{black}{#1}}

\definecolor{darkgreen}{RGB}{6,92,9}

\newcommand{\cmark}{\textcolor{darkgreen}{\ding{51}}}%
\newcommand{\xmark}{\textcolor{red}{\ding{55}}}%

\definecolor{codegreen}{rgb}{0,0.6,0}
\definecolor{codegray}{rgb}{0.5,0.5,0.5}
\definecolor{codepurple}{rgb}{0.58,0,0.82}
\definecolor{backcolour}{rgb}{1.0,1.00,1.00}

\lstdefinestyle{mystyle}{
    backgroundcolor=\color{backcolour},   
    commentstyle=\color{codegreen},
    keywordstyle=\color{magenta},
    numberstyle=\tiny\color{codegray},
    stringstyle=\color{codepurple},
    basicstyle=\ttfamily\tiny,
    breakatwhitespace=false,         
    breaklines=true,                 
    captionpos=b,                    
    keepspaces=true,                 
    numbers=left,                    
    numbersep=5pt,                  
    showspaces=false,                
    showstringspaces=false,
    showtabs=false,                  
    tabsize=2
}

\begin{document}

\title{\system: An Optimized Data System for Multi-Large-Model Deep Learning Workloads}

\author{Kabir Nagrecha}
\affiliation{%
	\institution{University of California, San Diego}
	\city{}
	\country{}
}
\email{knagrech@ucsd.edu}

\author{Arun Kumar}
\affiliation{%
	\institution{University of California, San Diego}
	\city{}
	\country{}
}
\email{akk018@ucsd.edu}

\begin{abstract}
Large models such as GPT-3 and ChatGPT have transformed deep learning (DL), powering applications that have captured the public's imagination. 
Such models must be trained on multiple GPUs due to their size and computational load, driving the development of a bevy of ``model parallelism'' techniques and tools.
Navigating such \textit{parallelism} choices, however, is a new burden for DL users such as data scientists, domain scientists, etc., who may lack the necessary systems knowhow. 
The need for \textit{model selection}, which leads to many models to train due to hyper-parameter tuning or layer-wise finetuning, compounds the situation with two more burdens: \textit{resource apportioning} and \textit{scheduling}. 
In this work, we unify these three burdens by formalizing them as a joint problem that we call \spase: \textsc{S}elect a \textsc{P}arallelism, \textsc{A}llocate resources, and \textsc{S}chedul\textsc{e}. 
We propose a new information system architecture to tackle the \spase~problem holistically, \red{exploiting the performance opportunities presented by joint optimization.}
We devise an extensible template for existing parallelism schemes and combine it with an automated empirical profiler for runtime estimation. We then formulate \spase~as an MILP. 
We find that direct use of an MILP-solver is significantly more effective than several baseline heuristics. 
We optimize the system runtime further with an introspective scheduling approach. 
We implement all these techniques into a new data system we call~\system. 
Experiments with benchmark DL workloads show that~\system~achieves 39-49\% lower model selection runtimes than current DL practice.
\end{abstract}

\maketitle

\pagestyle{\vldbpagestyle}
\begingroup\small\noindent\raggedright\textbf{PVLDB Reference Format:}\\
\vldbauthors. \vldbtitle. PVLDB, \vldbvolume(\vldbissue): \vldbpages, \vldbyear.\\
\href{https://doi.org/\vldbdoi}{doi:\vldbdoi}
\endgroup
\begingroup
\renewcommand\thefootnote{}\footnote{\noindent
This work is licensed under the Creative Commons BY-NC-ND 4.0 International License. Visit \url{https://creativecommons.org/licenses/by-nc-nd/4.0/} to view a copy of this license. For any use beyond those covered by this license, obtain permission by emailing \href{mailto:info@vldb.org}{info@vldb.org}. Copyright is held by the owner/author(s). Publication rights licensed to the VLDB Endowment. \\
\raggedright Proceedings of the VLDB Endowment, Vol. \vldbvolume, No. \vldbissue\ %
ISSN 2150-8097. \\
\href{https://doi.org/\vldbdoi}{doi:\vldbdoi} \\
}\addtocounter{footnote}{-1}\endgroup

\ifdefempty{\vldbavailabilityurl}{}{
\vspace{.3cm}
\begingroup\small\noindent\raggedright\textbf{PVLDB Artifact Availability:}\\
The source code, data, and/or other artifacts have been made available at \url{\vldbavailabilityurl}.
\endgroup
}

\section{Introduction}\label{sec:intro}

\begin{figure*}[t]
\includegraphics[width=0.9\textwidth]{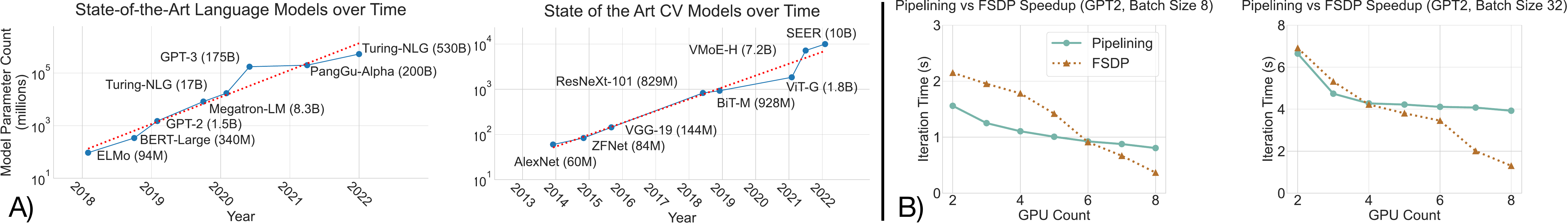}
\caption{(A) Trends of the sizes of some state-of-the-art DL models in NLP and CV (log scale), extrapolated from a similar figure in~\cite{megatron2019}. (B) Our empirically measured runtime crossovers between FSDP and pipeline parallelism, with knobs tuned per setting.}
\label{fig:open_panel}
\end{figure*}

Large-model deep learning (DL) is growing in adoption across many domains for data analytics over text, image, video, and even multimodal tabular data. 
Large language models (LLMs) now power popular applications like ChatGPT~\cite{chatgpt}. 
Such models~\cite{bert2018} have been ballooning in size, as Figure~\ref{fig:open_panel}(A) shows. 
For instance, the popular GPT-J~\cite{radford2019language,gpt-j} and ViT~\cite{vit2020} models need 10s of GBs of GPU memory and take days to train. 
This is often impractical for DL users in smaller companies, enterprises, and the domain sciences.
Thankfully, in most cases they need not train from scratch to benefit from large-model DL. 
They can download ``base'' models, pre-trained on large general datasets (e.g., Web-scraped text), from model hubs like HuggingFace~\cite{huggingface2019} and just ``finetune'' them on their (smaller) application-specific data~\cite{bommasani2022opportunities}. This enables companies to keep their application data in house. Recent market research reports that this form of large-model DL is rapidly growing~\cite{databricksreport}. 

While finetuning and customizing of base models has made large-model DL more tractable, end users of DL still face 3 systems-oriented headaches: (1) \textit{GPU memory} remains a bottleneck. Large-memory GPUs are expensive, and even public cloud vendors still ration them. (2) \textit{Multi-GPU parallelism} is needed but understanding the performance behaviors of complex large-model parallelism techniques is difficult for DL users; and (3) \textit{Model selection}, which involves tuning hyper-parameters, model layers, etc., only amplifies the computational load.

\vspace{1mm}
\textit{Overall, large-model DL, including finetuning, is still painful for regular DL users, hurting usability and raising runtimes and costs, especially in pay-as-you-go clouds.}
\vspace{3mm}

\noindent \textbf{Case Study:} Consider a data scientist, Alice, building an SQL autocomplete tool to help database users at her company. 
She has a (private) query log that contains her company's database schemas, common predicates, etc.
She downloads two LLMs from HuggingFace --- GPT-2 and GPT-J --- both of which are known to offer strong results for textual prediction tasks~\cite{radford2019language,gpt-j}.
She finetunes multiple instances on her dataset, comparing different batch sizes and learning rates to raise accuracy.
She uses an AWS instance with 8 A100 GPUs. 
She launches the DL tuning jobs in parallel, assigning one GPU each.
Alas, all of them crash with out-of-memory (OOM) errors.
She is now forced to pick a large-model scaling/parallelism technique and assign multiple GPUs to each job.
But to do so she must answer 3 intertwined systems-oriented questions: 
(1) Which parallelism technique to use for each model? 
(2) How many GPUs to assign to each model? 
(3) How to orchestrate such complex parallel execution for model selection workloads?

\vspace{3mm}
\textit{In this paper, we tackle precisely those 3 practical questions in a unified way to make it easier, faster, and cheaper for regular DL users like Alice to benefit from such state-of-the-art large DL models.}

\subsection{Prior Art and Their Limitations}\label{sec:prior_art}

We start by first explaining why prior art for large-model and parallel DL systems is insufficient to tackle the problem.  
Table~\ref{tb:prior_art} lists a conceptual comparison of our setting with prior art on several key aspects. 
Section~\ref{sec:related_work} discusses related work in greater detail.

\textit{(1) Which parallelism technique to use for each model?} 
There are a multitude of techniques in the ML systems world to parallelize/scale large models across GPUs. 
Some common techniques are: sharding the model, spilling shards to DRAM~\cite{swapadvisor2021,meng2017training}, pipeline parallelism as in GPipe~\cite{gpipe2018}, fully-sharded data-parallel (FSDP) as in PyTorch~\cite{torchfsdp2021} and ZeRO~\cite{zero2019}, hand-crafted hybrids as in Megatron~\cite{megatron2019}, as well as general hybrid-parallel approaches such as Unity~\cite{unger2022unity,flexflow2018} and Alpa~\cite{alpa2022}. 
But no technique dominates all others in all cases. 
Relative efficiency depends on a complex mix of factors: hardware, DL architecture specifics, even batch size for stochastic gradient descent (SGD).
Figure~\ref{fig:open_panel}(B) shows two empirical results on real workloads to prove our point. 
Even between just pipelining and FSDP, complex crossovers arise as GPU counts and batch sizes change. 
Furthermore, many techniques expose knobs that affect runtimes in hard-to-predict ways~\cite{terapipe2021}, e.g., pipelining requires tuning partitions and ``microbatch'' sizes, while FSDP requires tuning offloading and checkpointing decisions. \textit{Thus, we need to automate parallelism technique selection for large-model DL training.}

\begin{table*}[t]
\centering
\caption{Overview of prior art. Column desiderata are described in Sections~\ref{sec:prior_art} and~\ref{sec:desiderata}.}
\vspace{-3mm}
\begin{tabularx}{\textwidth}{ X P{2.5cm} P{1cm} P{2cm} P{1.5cm} P{1.5cm} P{3.5cm}} 
&  & Fidelity & Multi-Model & Resource \newline Allocation & Parallelism \newline Selection & Out-of-the-Box \newline Large Model Support \\
 \toprule 
\multirow{3}{*}{\textbf{\shortstack{Hybrid Parallelism}}} & Alpa~\cite{alpa2022} & \cmark & \xmark & \xmark & \cmark (limited) & \cmark  \\ 
 							       & FlexFlow~\cite{flexflow2018} & \cmark & \xmark & \xmark & \cmark (limited) & \xmark  \\ 
                                                                   & Unity~\cite{unger2022unity} & \cmark & \xmark & \xmark & \cmark (limited) & \cmark  \\  
                                                                                \midrule
\textbf{\shortstack{Performance Evaluation}} & Paleo~\cite{qi2016paleo} & \cmark & \xmark & \xmark & \cmark (limited) & \xmark  \\ 
									    \midrule
\multirow{2}{*}{\textbf{\shortstack{Model Selection}}}  & Cerebro~\cite{kumar2021cerebro} & \cmark & \cmark & \xmark & \xmark & \xmark  \\ 
									       & ASHA~\cite{asha2018} & \cmark & \cmark & \cmark & \xmark & \xmark  \\ 
									       \midrule
\multirow{3}{*}{\textbf{Scheduling}}  & Gandiva~\cite{xiao2018gandiva} & \cmark & \cmark & \xmark & \xmark & \xmark  \\ 
 									      & Antman~\cite{xiao2020antman} & \cmark & \cmark & \xmark & \xmark & \xmark  \\ 
 									      & Tiresias~\cite{gu2019tiresias} & \cmark & \cmark & \xmark & \xmark & \xmark  \\ 
									      \midrule
\multirow{2}{*}{\textbf{\shortstack{Resource Allocation}}} & Pollux~\cite{qiao2021pollux} & \xmark & \cmark & \cmark & \xmark & \xmark  \\ 
									    & Optimus~\cite{peng2018optimus} & \xmark & \cmark & \cmark & \xmark & \xmark  \\ 
									    \midrule
\textbf{\spase} & \system~(ours) & \cmark & \cmark & \cmark & \cmark & \cmark  \\ 
 
 \label{tb:prior_art}
\end{tabularx}
 \vspace{-8mm}
\end{table*}

\textit{(2) How many GPUs to assign to each model?} 
Many DL practitioners use fixed clusters or have bounded resource budgets. 
So, they are either given (or decide) up front the number of GPUs to use.
But in multi-model settings like model selection, there is more flexibility on apportioning GPUs across models. 
The naive approach of running models one after another using all GPUs is sub-optimal as it \textit{reduces model selection throughput} and adding more GPUs per model yields diminishing returns. 
Alas, the scaling behaviors of large-model parallelism techniques are not linear and often hard to predict, as Figure~\ref{fig:open_panel}(B) shows.
Prior art has studied data-parallel resource allocation (e.g., Pollux~\cite{qiao2021pollux} and Optimus~\cite{peng2018optimus}) and model selection optimization (e.g., Cerebro~\cite{kumar2021cerebro} and ASHA~\cite{asha2018}).
But none of them target large-model DL, which alters the cost-benefit tradeoffs of GPU apportioning in new ways due to interplay with parallelism selection and complex scaling behaviors. 
\textit{Thus, we must automate GPU apportioning for large-model model selection.}

\textit{(3) How to orchestrate such complex parallel execution for model selection?}
This is a scheduling question, i.e., deciding which jobs to run when.
Two naive approaches are to run models in a random order or to use a generic task scheduler.
Both can lead to GPU idling due to a lack of awareness of how long models actually run.
Prior art has studied runtime-aware DL scheduling, e.g., Gandiva~\cite{xiao2018gandiva} and Tiresias~\cite{gu2019tiresias}, but none target large-model DL. 
The complex interplay of parallelism selection and GPU apportionment can affect runtimes in a way that alters the tradeoffs of scheduling.
The model selection setting adds more considerations: we must optimize end-to-end \textit{makespan} rather than just a throughput objective~\cite{qiao2021pollux,peng2018optimus}. 
Specific desiderata must be met: \textit{fidelity} on ML accuracy and \textit{generality} on specification. We expand on these in Section~\ref{sec:desiderata}.

\textit{Overall, there is a pressing need for a unified and automated way to tackle these 3 systems concerns of model selection on large models: select parallelism technique per model, apportion GPUs per model, and schedule them all on a given cluster. No prior art --- including all those described in Table~\ref{tb:prior_art} --- can address this novel setting that has emerged with the rise of large-model DL. We call this new joint problem \spase: Select Parallelism, Apportion resources, and SchedulE.}

\subsection{System Desiderata}\label{sec:desiderata}

To help democratize large-model DL and ease practical adoption, we seek a data system that tackles \spase~with the following desiderata: 

\textbf{(1) Extensibility on parallelism selection.}
Given the variety of large-model parallelism techniques (henceforth called ``parallelisms'' for brevity), the system must support and select over multiple parallelisms and also make it easy for users to add new parallelisms in the future (e.g. for model-technique codesign~\cite{dlrm2019,megatron2019,fedus2021switch}). Without support for user extension, parallelism selectors/hybridizers are limited in scope, as noted in Table~\ref{tb:prior_art}.

\textbf{(2) Non-disruptive integration with DL tools.}
The system must natively support popular DL tools such as PyTorch~\cite{torchddp2020} and TensorFlow~\cite{tensorflow2015-whitepaper} without modifying their internals. This can offer backward compatibility as those tools evolve.

\textbf{(3) Generality on multi-model specification.}
The system should support multiple model selection APIs, e.g., grid/random search or AutoML heuristics.
We assume the system is given a set of model training jobs with known epoch counts. 
Evolving workloads can be supported by running all models one epoch at a time.

\textbf{(4) Fidelity on ML accuracy.}
The system must not deliberately alter ML accuracy when applying system optimizations. 
Approximations such as altering the model, training algorithm, or workload parameters are out of scope because they can confound users.

\begin{figure}[t]
\includegraphics[width=0.48\textwidth]{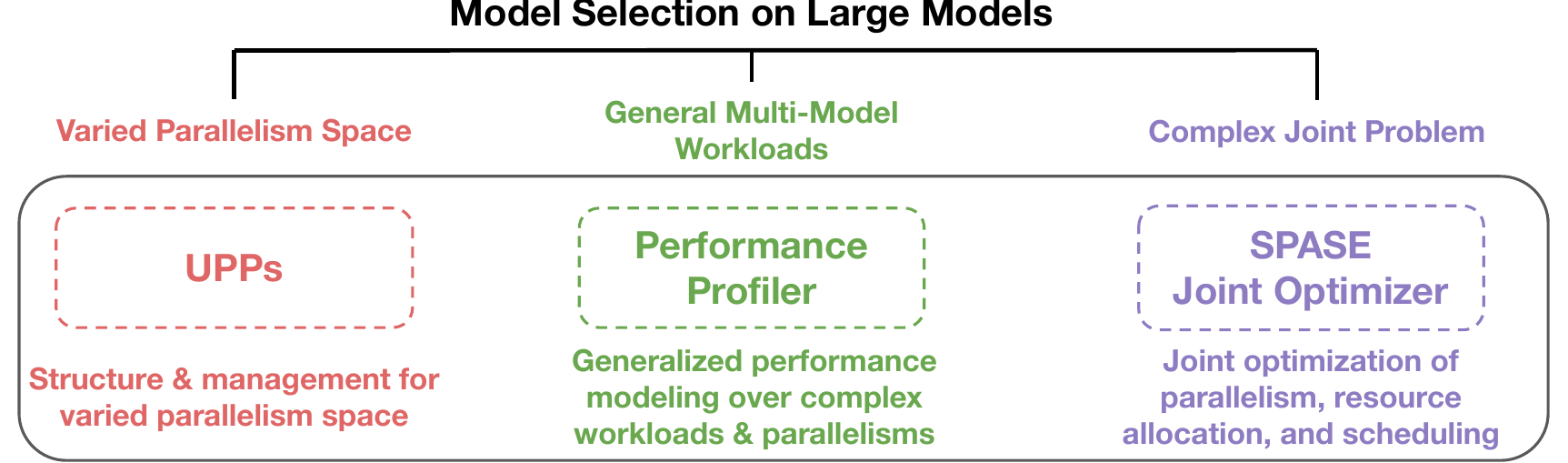}
\vspace{-5mm}
\caption{Overview of how \system's components tackle the SPASE problem for multi-large-model DL workloads.}
\label{fig:highlevel}
\vspace{-3mm}
\end{figure}

\subsection{Our Proposed Approach}

To meet all of the above desiderata, we design a new information system architecture to tackle \spase~that is inspired by some techniques in database systems. We call our system \system. 
Our current focus is on the common fixed-cluster setting rather than autoscaling~\cite{roy2011efficient}.
As Figure~\ref{fig:highlevel} shows, our approach is three-pronged:

\textbf{(1) Parallelism Selection and UPPs.}
We translate high-level (``logical'') model training specifications into optimized ``physical'' parallel execution plans based on instance details, inspired by physical operator selection in RDBMSs, e.g., selecting hash-join vs.~sort-merge join for a given join operation. 
To meet the first desideratum of extensibility, we introduce the abstraction of User-Pluggable Parallelisms (UPPs).
UPPs can be used to specify existing parallelisms in standard DL tool code, or enable users to add new parallelisms as blackboxes for \system~to use. This also ensures the second desideratum of non-disruptive integration. 
We create a default UPP library in \system~to support 4 major existing parallelisms: pipelining, spilling, distributed data parallelism (DDP), and FSDP. 
Each UPP can support knob-autotuning, similar to auto-tuning of physical configuration parameters of a data management system~\cite{ottertune,herodotou2011starfish}.

\textbf{(2) Performance Profiling.}
To apportion GPUs and select parallelisms in a way that ensures the fourth desideratum, we need accurate estimates of job runtimes \textit{as is}. 
We exploit a basic property of SGD: since minibatch size is fixed within an epoch, we can typically project epoch times accurately from runtime averages over a few minibatch iterations. 
\red{This is similar to prior works (e.g. the Clockwork inference system~\cite{clockwork2020}) that exploit the deterministic and predictable performance behaviors displayed by DNNs to proactively plan out high-quality execution schemes.}
Coupled with the offline nature of model selection, we can create a general and effective solution: profile all jobs using the full ``grid'' of options for both GPU counts and parallelisms based on only a few minibatches.  
The overhead of this approach is affordable due to the long runtimes of actual DL training.
This also ensures our second and third desiderata as all DL tools offer data sampling APIs that we can just use on top of the user-given model specifications.
Of course, we use the full training data for the actual DL jobs to ensure the fourth desideratum.

\textbf{(3) Joint Optimization and Scheduling.} 
Given the above system design choices, we can now tackle \spase~using joint optimization.
We formalize this problem as a mixed-integer linear program (MILP). 
Using realistic runtime estimates, we perform a simulation study to compare an MILP solver (we use Gurobi~\cite{gurobi}) to a handful of strong scheduling heuristics. 
The solver yields the best results overall even with a timeout. 
Thus, we adopt it in \system~as our \spase~optimizer. 
Actual model training, not the optimizer, heavily dominates overall runtimes in DL workloads, so we view this design decision as reasonable because it ensures \textit{both efficiency and simplicity}, easing system maintenance and adoption. 
Finally, we augment our Optimizer with an ``introspective'' scheduling extension known in prior art to further raise resource utilization.

We intentionally design \system~to be a simple and intuitive system to tackle \spase~in a way that can help ease practical adoption. Figure~\ref{fig:sys_architecture} in Section 3 shows our system architecture. 
\system~is implemented in Python and exposes high-level APIs for (offline) specification of UPPs and model selection APIs for actual DL training usage. 
Under the hood, \system~has 4 components: Parallelism Plan Enumerator, Performance Profiler, Joint Optimizer, and Executor. 
The runtime layer builds on top of the APIs of the massively task-parallel execution engine Ray~\cite{ray2017} for lower level machine resource management, e.g., placing jobs on GPUs, as well as to parallelize our profiling runs. 
Using two benchmark large-model workloads from DL practice, we evaluate \system~against several baselines, including an emulation of current practice of manual decisions on \spase. 
\system~reduces overall runtimes by 39\% to 49\%, which can yield proportionate cost savings on GPU clusters, especially in the cloud. We perform an ablation study to isolate the impacts of our optimizations. Finally, we evaluate \system's sensitivity to the sizes of models, workloads, and nodes.

\textbf{Novelty \& Contributions.} To the best of our knowledge, this is the first work to unify these three critical requirements of large-model DL workloads for end users: parallelism selection, resource apportioning, and scheduling. 
By casting the problem this way, we judiciously synthesize key system design lessons to craft a new information system architecture that can reduce user burden, runtimes, and costs via joint optimization in this important analytics setting. 
Overall, this paper makes the following contributions:

\begin{itemize}[leftmargin=*]

\item We formalize and study the unified \spase~problem, freeing end users of large-model DL from having to manually select and tune parallelisms, apportion GPUs, and schedule multi-jobs. 

\item We present \system, a new information system architecture to tackle \spase~that is also the first to holistically optimize parallelism selection and resource apportioning for multi-large-model DL. \system~employs a generalized profiler to estimate parallelism runtimes and an MILP solver for joint optimization. 

\item To enable generalized and extensible support for parallelisms, we create the abstraction of User-Defined-Parallelisms (UPPs). UPPs can be used to specify parallelisms as blackboxes in \system. 


\item We perform an extensive empirical evaluation of \system~on two benchmark large-model DL workloads. \system~reduces model selection runtimes by up to 49\% in some cases. \red{We make our code publicly available on GitHub~\footnote{https://github.com/knagrecha/saturn}.}

\end{itemize}


\section{Background and Preliminaries}\label{sec:preliminaries}
\red{We provide a brief background on parallelization techniques to describe the fundamentals relevant to our problem space. 
For the interested reader, we provide a broader overview of the ML Systems space in the Appendix of our tech report~\cite{techreport}.}

\noindent \textbf{Multi-GPU parallelism} is now common in large-model DL training~\cite{dnnCluster2019}. 
Several parallelization schemes already exist, and researchers continue to routinely devise and propose new techniques. 
A comprehensive review of all such approaches is out of scope for this paper; we refer interested readers to the relevant surveys~\cite{researchExam,lowmemory2019}. 
\red{Instead, we only highlight a few common approaches here for reference. We also mention the tunable knobs for each parallelism that complicate scaling behaviors and theoretical performance analyses.} 

\textit{Data Parallelism} replicates a given DL model across multiple accelerators. 
Each is fed a different minibatch partition for parallel processing. Replica synchronization can be done in two ways --- either via a central parent server, for Parameter Server (PS)-style data parallelism~\cite{li2013parameter,Renz_Wieland_2022},
or through peer-to-peer communication, for all-reduce data parallelism~\cite{sergeev2018horovod,torchddp2020} with synchronization at SGD boundaries.

\textit{Model Parallelism} partitions the \textit{model} rather than the data. 
The model graph is sharded and partitioned over GPUs to distribute the memory footprint. 
\red{The speedup potential of model parallelism depends on the partitioning scheme and model architecture.} 
Hand-crafted, architecture-specific approaches can perform well~\cite{megatronlmblog2020}, while simple and generic partitioning schemes tend to be slower~\cite{mpms2021}.

\red{\textit{Pipelining~\cite{yang2020pipemare,gpipe2018,torchgpipe2020,terapipe2021} \& Fully-Sharded Data Parallelism (FSDP)~\cite{zero2019,torchddp2020}} are more \red{advanced hybridizations of model parallelism with data parallelism.
Each presents its own tradeoffs and optimization knobs (e.g. ``microbatches'' for pipelining~\cite{terapipe2021}, and ``offloading'' and ``checkpointing''~\cite{checkpointing2016} for FSDP).} For brevity, we elaborate on the specifics of these techniques in the Appendix of our tech report~\cite{techreport}.}

\textit{Spilling} is not a parallelism technique in itself but is often used in combination with a parallelism technique to reduce GPU memory pressure. It swaps model shards between GPU memory and DRAM for piece-wise GPU-accelerated execution~\cite{mpms2021,FairScale2021}. 
This adds DRAM-GPU communication overheads, but it can enable large models to be trained with even just one GPU. 
Spilling exposes a \textit{partition count} knob, to select the number of DRAM spills during execution.

\noindent \textbf{Model selection} is the process of training and comparing model configurations. 
Two popular procedures are \textit{grid search}, in which all combinations of sets of values of hyper-parameters (e.g., batch size, learning rate) are used, and \textit{random search}~\cite{bergstra2012random}, in which random hyper-parameter combinations from given intervals are used. Early stopping can reduce the set of configurations during training~\cite{asha2018,hyperband,hyperdrive2017}. 
The high resource demands of model selection on large models can sometimes force a DL user to settle for a smaller search space, but this risks missing out on higher accuracy~\cite{dl-book,kumar2016model}. 
Faster execution of such workloads empowers users to run larger searches, in turn helping accuracy. 
\red{Many users expect \textit{fidelity} in this setting, as we explain in Section~\ref{sec:desiderata}.}

\section{System Overview}\label{sec:overview}
\begin{figure}[t]
\includegraphics[width=0.9\columnwidth]{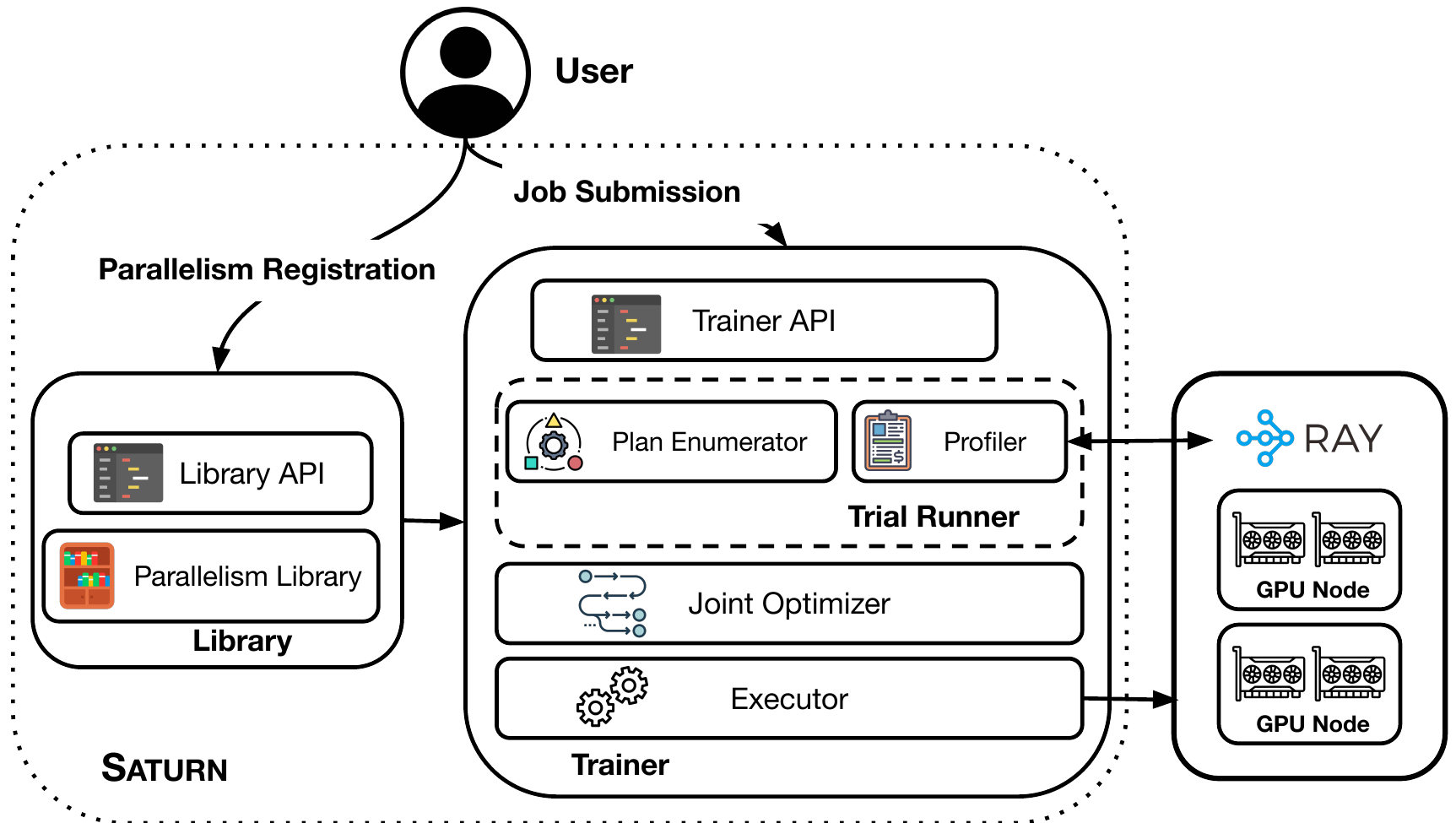}
\caption{System architecture of~\system~and the interactions between the components.}
\label{fig:sys_architecture}
\end{figure}

We now describe \system's architecture that meets the desiderata in Section 1.2. 
\system~has 4 main modules, as Figure~\ref{fig:sys_architecture} shows. 
For workload specification, it exposes a high-level API and the Parallelism Library. 
The Trial Runner handles runtime estimation. The Joint Optimizer and Executor tackle the \spase~problem. 
\system~uses Ray~\cite{ray2017}'s low-level APIs as the runtime layer that places jobs on GPUs. 
Next, we describe each of \system's~components.

\subsection{Workload Specification}\label{sec:workload_spec}
The first phase, workload specification, is handled by our API and the Parallelism Library component.

\textbf{API.}
~\system's API provides an easy-to-use interface for both registering parallelisms (for developers) and submitting large-model training jobs (for end users of DL). 
We now provide a brief overview; due to space constraints, we provide the full example pseudocode in the technical report~\cite{techreport}. 
There are two parts to the API: the Library API and the Trainer API.
Users create ``Tasks'' through the Trainer API by specifying functions for model initialization and data loading, along with any hyper-parameters. This is sufficiently general to cover most model selection workloads. 
Listing~\ref{lst:task_spec} illustrates.

\begin{lstlisting}[language=Python, label={lst:task_spec},caption=Specifying tasks through \system's API.,frame=single]
from saturn.trainer import Task, HParams, execute, profile

t_1=Task(get_model,get_data,HParams(lr=1e-3,epochs=5,optim=SGD))
t_2=Task(get_model,get_data,HParams(lr=3e-3,epochs=5,optim=SGD))
\end{lstlisting}

\noindent Training procedures are defined by ``User-Pluggable Parallelisms'' (UPPs), which implement the parallel execution approach for SGD. 
These parallelisms can be registered with our Library by a developer (e.g., ML engineer) or a system-savvy end user of DL. The registration process is shown in Listing~\ref{lst:technique_registration}.

\begin{lstlisting}[language=Python, label={lst:technique_registration},caption=Parallelism registration.,frame=single]
from saturn.library import register

register("parallelism-a", ParallelismA)
register("parallelism-b", ParallelismB)
\end{lstlisting}

\noindent Once all parallelisms and tasks are specified, DL users can invoke the Trial Runner to produce runtime estimates in a single line of code, followed by invoking the whole training execution in another single line of code. 
Listing~\ref{lst:execution} illustrates these. 

\begin{lstlisting}[language=Python, label={lst:execution},caption=Profiling and execution invocations.,frame=single]
profile([t_1, t_2, t_3])
execute([t_1, t_2, t_3])
\end{lstlisting}

%
%
%
%
%
%

\textbf{Parallelism Library.} 
The design of this library is inspired by functional frameworks, user-defined function templates in RDBMSs, and DL model hubs~\cite{huggingface2019}. 
We follow a define-once, use-anywhere design, wherein registered UPPs can be reused across models, execution sessions, and even different cluster users. 
This is achieved by managing library-registered parallelisms as a database of code files. 
The Library allows developers to register new parallelisms by implementing an abstract skeleton, shown in Listing~\ref{lst:technique_spec}.

\begin{lstlisting}[language=Python, label={lst:technique_spec},caption=Parallelism specification skeleton.,frame=single] 
class BaseParallelism:
	def search(task:Task,gpus:List[int])->Dict,float:
		pass
	def execute(task:Task,gpus:List[int],knobs:Dict)->None:
		pass
\end{lstlisting}

\noindent The \textit{search} function should use the task and GPUs to provide (1) execution parameters (e.g., microbatch count, partition count) and (2) a runtime estimate. 
Knob-optimization can also optionally be tackled here. 
Failed searches (e.g., OOMs) can be handled by returning null values.
The \textit{execute} function trains the provided task to completion using the allotted GPUs. 
It also uses any execution parameters produced during the search phase to optimize execution.

Developers can implement a UPP with standard DL tool code (e.g., TensorFlow or PyTorch) without restrictions. This enables easy integration of pre-existing parallelisms. 
Indeed, we validate that functionality by adding 4 major parallelisms in our default Parallelism Library: DDP~\cite{torchddp2020}, GPipe-style pipeline parallelism~\cite{torchgpipe2020}, FSDP, and model spilling via the FairScale package~\cite{FairScale2021}.
These out-of-the-box parallelisms in \system~are maximally general in that they can be automatically applied to any DL model supported by them. Implementing UPPs for each took $100-250$ lines of Python code. Once defined, UPPs can be registered with the Library under a user-set name (e.g. ``pytorch-ddp'').

Our design can help developers retain a familiar environment without low-level code changes or extraneous workflows to, say, translate their parallelism implementation into a new configuration file format, a custom domain specific language, etc. 
Our Parallelism Library serves as an organized roster for registering and using large-model DL parallelisms. 
While it is a key part of \system, it can potentially also be useful as its own standalone tool. 

\subsection{Performance Estimation}
The Trial Runner estimates the runtime performance of models with different parallelisms and GPU apportionments. The Trial Runner is \textit{not} a parallelism selector: it simply generates the statistics needed to solve \spase. 
It is our empirical substitute for the complex parallelism-specific theoretical models used in prior art~\cite{qiao2021pollux,peng2018optimus}. Such empirical profiling helps ``future proof'' \system~to an extent: by not tightly coupling \system~to specific parallelisms' theoretical models, we can directly support future DL tool compilers and/or accelerator hardware as they evolve. 
As we highlight in Section~\ref{sec:desiderata}, extensibility is one of our key desiderata.
The Trial Runner has two submodules: Plan Enumerator and Profiler. 

\textbf{Plan Enumerator.} 
This sub-module constructs a ``grid'' across all supported parallelisms and GPU apportionment levels for each model. That represents the space of ``physical plans'' for every model that will then be profiled to obtain runtime performance estimates. 

\textbf{Profiler.} This sub-module takes the outputs of the Plan Enumerator to produce runtime estimates for the optimization phase. 
We exploit a property of SGD: since it is iterative and consistent, we can accurately extrapolate epoch runtimes from averaged performance over a just few minibatches~\red{\cite{clockwork2020}}. 
We use Ray to parallelize these profiling runs and reduce the Profiler's runtime. 
In our experiments, profiling 12 multi-billion-parameter models for 4 parallelisms took < 30min. 
This overhead is affordable because the actual DL model selection, on the full training data, can take hours or even days.



\subsection{Joint Optimizer and Executor}
We now use the Trial Runner's statistics to tackle the \spase~problem in a unified manner via holistic optimization. 

\textbf{Joint Optimizer.} 
The Joint Optimizer is invoked transparently when the user invokes the \textit{execute} function. 
It uses the runtime estimates produced by the Trial Runner and cluster details to produce a full execution plan. 
This plan bakes in all of \textit{parallelism selection}, \textit{GPU apportionment}, and \textit{schedule construction}.
To construct the plan, the Joint Optimizer automatically determines the following for all model configurations given by the user: (1) which parallelism to use, (2) how many GPUs to give it, and (3) when to schedule it. 

Our Optimizer is implemented in two layers. First, an MILP solver to produce makespan-optimized execution plans. 
Second, an introspective, round-based resolver that runs on top of the MILP solver to support dynamic reallocation.
Section~\ref{sec:optimizations} goes into the technical details of the MILP, why we chose to use an MILP solver instead of heuristics, and additional techniques in the Joint Optimizer.

\textbf{Executor.} This module handles the running of the full execution plan generated by the Joint Optimizer. The Executor runs on top of the lower level APIs of Ray to leverage its task-parallel processing.  
By default, Ray uses its own task scheduler, and swapping that out for a custom scheduler is challenging. 
So, for the Executor we implement our plan \textit{over} Ray's scheduler. 
We achieve this by ``tainting'' Ray-owned GPUs so that they can only be used by the corresponding jobs from our pre-calculated schedule. 
Thus, the Executor \red{ensures that Ray's scheduler cannot deviate from our \spase~solution.
This scheme lets us faithfully recreate the optimizer-designed plan without overheads or induced inefficiencies, even though the design goes beyond Ray's intended usage.}

\subsection{Current Limitations}
\system~supports both single-node and multi-node training across different models, but in the current version we focus on the case where each model fits in aggregate node memory (i.e., total GPU memory + DRAM).
\red{Since we focus on the large-model case, we do not consider GPU multi-tenancy (e.g., as in ModelBatch~\cite{narayanan2018accelerating}).} 
We also focus on the homogeneous GPU cluster setting and leave to future work adding support for heterogeneous hardware clusters, hardware type selection, and elastic provisioning (e.g., like in~\cite{gavel2020,heteroelastic}). 
Anecdotally, we find that many DL users in domain sciences and enterprises do indeed fit this setting. 
Furthermore, many of the parallelisms in our existing Library do not yet support cross-node training for a single model out-of-the-box. So, we defer support to a future extension as those parallelisms evolve. 
Despite these assumptions, \system~can already train 10B+ parameter models on even just one node. 
These limitations can be mitigated in the future as follows: (1) adjust the MILP in Section~\ref{sec:optimizations} for hardware selection, (2) give the Trial Runner a larger space to explore, and (3) add multi-node parallelisms to the Library~\cite{zhang2017poseidon}. 

\red{Two other relevant extensions are support for autoscaling support and elastic re-configurations of jobs mid-execution. An obvious and straightforward way to incorporate these extensions would be to submit workloads to \system~one-epoch-at-a-time, then induce environment/workload changes at a higher level, in between \system's invocations. Future work could look to support more fine-grained integrations, e.g., where \system~controls the autoscaling decisions. We discuss some possible adaptation points in Section~\ref{sec:dynamic_rescaling}, but leave these extensions to future work.}
\section{SPASE Joint Optimizer}\label{sec:optimizations}

We now describe the \spase~problem and dive into our MILP formalization. 
Using a simulation study, we evaluate an MILP solver (Gurobi~\cite{gurobi}) against baselines and heuristics from standard practice and prior art. 
We explain our introspective mechanism that enables \system~to adaptively reassess its MILP solution over time.

\subsection{Problem Basics}\label{sec:resource_scheduling}

\spase~unifies parallelism selection, resource allocation, and schedule construction. 
Typical schedulers can set task start times, while resource schedulers can select a GPU apportionment as well. 
But with \spase, our joint optimizer must consider a third performance-critical dimension: select the parallelism to use for each model on the allotted GPUs. 
To the best of our knowledge, ours is the first work to unify and tackle this joint problem. 

In model selection workloads, it is common for all jobs to be given up front. So, we focus on that setting. 
Using the Trial Runner module, we generate the necessary runtime statistics for all given jobs. 
But even with that information, the joint problem is intractable; prior work on network bandwidth distribution~\cite{networkScheduling2014} has shown that even the basic resource allocation problem is NP-hard. 
\spase~is a more complex version of that problem that also handles parallelism selection and makespan-optimized scheduling; so it is also NP-hard. 
Brute-forcing the search space is also impractical due to its sheer size. 
The number of schedule orderings alone grows super-exponentially with the number of jobs~\cite{tucker1994applied}. 
As such, solving it optimally is ruled out.
Thus, we choose to formulate \spase~as an MILP and use an industrial-strength MILP solver (Gurobi~\cite{gurobi}) to leverage its time-tested optimization power. 
Later, in Section~\ref{sec:sched_eval}, we justify this decision further using a simulation study. We find that the MILP solver significantly and consistently outperforms known baselines and strong heuristics despite its time limit. 
\red{We rely on Gurobi's sophisticated techniques to avoid pitfalls such as poor local optima~\cite{gurobibranch} in this highly non-convex optimization space. 
Even if the solver does only reach a local optimum, the solution should be of reasonably high quality. We describe and evaluate these risks further towards the end of Section~\ref{sec:milp_baselines} and the Appendix of our tech report~\cite{techreport}.}
To the best of our knowledge, ours is the first MILP formulation to unify DL parallelism selection, resource allocation, and scheduling. Not only does it enable us to state the problem with mathematical precision, it also enables us to explore the problem space's intricacies via the simulation study.

\begin{table}[t]\centering
\caption{MILP Notation used in Section~\ref{sec:milp}}
\label{table:milp_notation}
\vspace{-4mm}
\begin{tabular}{@{}l p{6.7cm}@{}}
\toprule
\multicolumn{2}{c}{\textbf{Inputs to the MILP}} \\
\midrule
Symbol & Description \\
\midrule
$N$ & List of nodes available for execution. \\
$T$ & List of input training tasks. \\
$U$ & Large integer value used to enforce conditional constraints. \\
$GPU_n$ & The number of GPUs available on node $n$. \\
$S_{t}$ & Number of configurations available to task $t$. A configuration consists of both a parallelism and a GPU allocation.\\
$G_{t} \in {\mathbb{Z}^+}^{S_{t}}$ & Variable length list of requested GPU counts for each configuration of task $t$. \\
$R_{t}\in {\mathbb{R}^+}^{S_{t}}$ & Variable length list of estimated runtimes for each configuration of task $t$.\\
\midrule
\multicolumn{2}{c}{\textbf{MILP Selected Variables}} \\
\midrule
Symbol & Description \\
\midrule
$C$ & Execution schedule makespan. \\
$B_{t} \in {0, 1}^{S_{t}}$ & Variable length list of binary variables indicating whether task $t$ uses the corresponding configuration from $S_{t}$.\\
$O_{t,n} \in {0,1}$ & Binary indicator of whether task $t$ ran on node $n$.\\
$P_{t,n,g} \in {0,1}$ & Binary indicator of whether task $t$ ran on GPU $g$ of node $n$.\\
$A_{t1, t2} \in {0,1}$ & Binary indicator of whether task $t1$ ran before task $t2$. If $A_{t1,t2}$ is 1, $t2$ must have run after $t1$.\\
$I_{t,n,g} \in {\mathbb{R}^+}$ & Start time of task $t$ on GPU $g$ of node $n$. \\

\bottomrule
\end{tabular}
\vspace{-6mm}
\end{table}

\subsection{MILP Formulation}\label{sec:milp}

\begin{figure*}[t]
\includegraphics[width=0.95\textwidth]{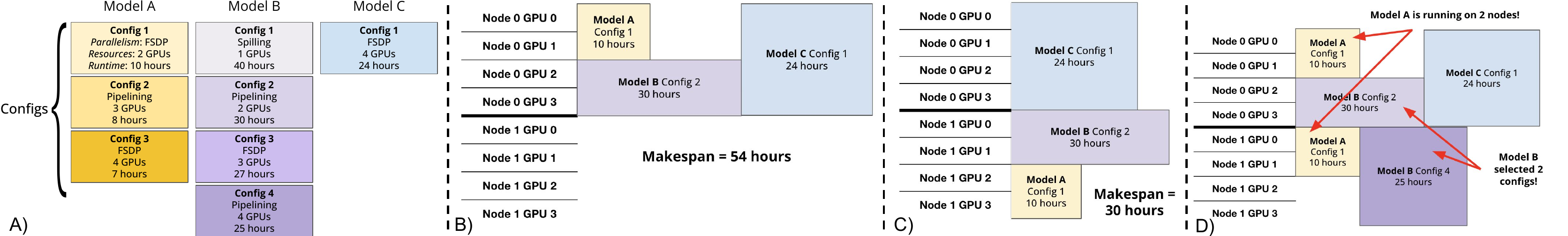}
\vspace{-3mm}
\caption{\red{(A) depicts the configs (i.e., variables $G$ \& $R$) used throughout our examples; (B) illustrates a feasible but suboptimal \spase~solution and the corresponding makespan; (C) illustrates an optimal \spase~solution; (D) illustrates violations of the constraints in Equation~\ref{eqn:binary_constraints_a}. The remainder of the constraints are illustrated in the Appendix of our tech report~\cite{techreport}.}}
\label{fig:milp_illustration}
\vspace{-4mm}
\end{figure*}

\textbf{Inputs.} Our MILP input consists of a full grid of models, their valid configurations, as well as the corresponding runtime estimates~\red{generated by the Trial Runner}.
Table~\ref{table:milp_notation} lists our notation, \red{and Figure~\ref{fig:milp_illustration}(A) illustrates an example}. 
As noted in Section~\ref{sec:overview}, our empirical runtime estimates already bake in the communication overheads of each parallelism. 

\textbf{Summary.} To summarize the MILP's function in plain-English: we ask the solver to assign to each task: (1) GPU IDs with associated node IDs, (2) an execution configuration (determining the parallelism and resource apportionment), and (3) a float start time.
Each task should only be assigned one node and one configuration, and the number of GPUs assigned should agree with the specifications of the chosen configuration.
The task should not block \textit{any} GPUs on a node it is not using. The start time for a given task should align all assigned GPUs (i.e., gang scheduling), and the assigned start times should not cause task overlaps on the same GPUs.
Ultimately, the solution should minimize the makespan.

\textbf{Formulation.} We now go into each constraint in depth. To make the formulation easier to comprehend, we illustrate our constraints using a running example workload in Figure~\ref{fig:milp_illustration} (continued in the Appendix of our tech report~\cite{techreport}). The figures are purely demonstrative, and do not represent a realistic model selection job.

\begin{align}
\vspace{-3mm}
	\label{eqn:milp_obj}
	\texttt{Objective:} ~\quad \displaystyle\min_{B,O,P,A,I} {C}
\vspace{-0.5mm}
\end{align}

We now define the constraints.
Equation~\ref{eqn:makespan} defines the makespan; it is the latest task's start time plus the runtime of that task's selected configuration.
\red{Figure~\ref{fig:milp_illustration}(B) \& (C) illustrate some example \spase~solutions and their corresponding makespans.}

\begin{align}
\vspace{-3mm}
\label{eqn:makespan}
\begin{split}
C \geq I_{t,n,g} + R_{t,s} - U \times (1 - B_{t,s}) \\ \forall s \in S_{t} \forall t \in T, \forall n \in N, \forall g \in G
\end{split}
\vspace{-0.5mm}
\end{align}

Next, \red{for each task, there should only be one selected configuration and only one selected node}.
\red{Figure~\ref{fig:milp_illustration}(D) illustrates this constraint.}

\begin{align}
\vspace{-3mm}
\label{eqn:binary_constraints_a}
\begin{split}
\sum_{x \in B_{t}} x = 1; \sum_{y \in O_{t}} y = 1
\end{split}
\vspace{-0.5mm}
\end{align}

Next, we enforce the GPU requests of the solver onto the execution schedule. \red{Each task must be assigned the number of GPUs corresponding to its selected configuration.
Since direct equality comparisons are not possible in an MILP formulation, Equations~\ref{eqn:gpu_allocation_a} and ~\ref{eqn:gpu_allocation_b} in combination ensure this constraint
by enforcing both $\leq$ and $\geq$ inequalities.}

\begin{align}
\label{eqn:gpu_allocation_a}
\begin{split}
\sum_{t \in P_{t,n}} t \geq G_{t,s} - U \times (2 - O_{t,n} - B_{t,s}) \forall s \in S_{t}, \forall t \in T, \forall n \in N
\end{split}
\end{align}

\begin{align}
\label{eqn:gpu_allocation_b}
\begin{split}
\sum_{t \in P_{t,n}} t \leq G_{t,s} + U \times (2 - O_{t,n} - B_{t,s}) \forall s \in S_{t}, \forall t \in T, \forall n \in N
\end{split}
\end{align}

\red{We must also ensure that the task uses \textit{0} GPUs on any nodes it is not executing on. Equations~\ref{eqn:gpu_unallocation_a} and~\ref{eqn:gpu_unallocation_b} combine $\leq$ and $\geq$ inequalities to enforce this requirement.} 
\red{This constraint and all subsequent ones are illustrated with examples in the Appendix of our tech report~\cite{techreport}.}

\begin{align}
\label{eqn:gpu_unallocation_a}
\begin{split}
\sum_{t \in P_{t,n}} t~\red{\leq}~0 - U \times (O_{t,n} + B_{t,s}) \forall s \in S_{t}, \forall t \in T, \forall n \in N
\end{split}
\end{align}

\begin{align}
\label{eqn:gpu_unallocation_b}
\begin{split}
\sum_{t \in P_{t,n}} t~\red{\geq}~0 + U \times (O_{t,n} + B_{t,s}) \forall s \in S_{t}, \forall t \in T, \forall n \in N
\end{split}
\end{align}

Next we apply a \textit{gang scheduling} constraint, \red{i.e. for each task, all assigned GPUs must initiate processing simultaneously.}
Formulating this constraint is challenging --- we need consistency over a set of MILP-selected values,  on a set of MILP-selected indices, across an MILP-selected gang size. 
Our solution is to take a fixed start-time target --- the sum of MILP-selected start times over \textit{all GPUs}, divided by the number of allocated GPUs.
\red{By ensuring each selected time is thus equal to the \textit{average} of the times, the times must by definition be equal to one another.}
This constraint \red{also} naturally encourages the solver to fix start times on unused GPUs to 0 without explicit enforcement, \red{since non-zero values bloat the numerator of the left hand side}. 
Equations~\ref{eqn:gpu_gang_a} and~\ref{eqn:gpu_gang_b} \red{in combination} enforce this constraint.

\begin{align}
\label{eqn:gpu_gang_a}
\begin{split}
\frac{\sum_{x \in I_{t,n}} x}{G_{t, s}} \leq I_{t,n,g} + U \times (3 - P_{t,n,g} - B_{t,s}  - O_{t,n}) \\ \forall s \in S_{t}, \forall t \in T, \forall g \in GPU_{n}, \forall n \in N
\end{split}
\end{align}
	
\begin{align}
\label{eqn:gpu_gang_b}
\begin{split}
\frac{\sum_{x \in I_{t,n}} x}{G_{t, s}} \geq I_{t,n,g} - U \times (3 - P_{t,n,g} - B_{t,s}  - O_{t,n}) \\ \forall s \in S_{t}, \forall t \in T, \forall g \in GPU_{n}, \forall n \in N
\end{split}
\end{align}

Finally, we encode a task isolation constraint, so that no tasks overlap on the same GPU. 
Equation~\ref{eqn:task_isolation_a} applies if task $t1$ came before task $t2$, while equation~\ref{eqn:task_isolation_b}
guarantees no overlap if task $t1$ came after task $t2$. \red{Variable $A$ acts as a before-or-after selector, determining which constraint is relevant for each pair of tasks.}

\begin{align}
\label{eqn:task_isolation_a}
\begin{split}
I_{t1,n,g} \leq I_{t2,n,g} - R_{t,s} + U \times ((3 - P_{t1,n,g} - P_{t2,n,g}) - B_{t,s} + A _{t2, t1}) \\ \forall s \in S_{t}, \forall t1 \in T, \forall t2 \in (T - \{t1\}), \forall g \in GPU_{n}, \forall n \in N
\end{split}
\end{align}

\begin{align}
\label{eqn:task_isolation_b}
\begin{split}
I_{t1,n,g} \geq I_{t2,n,g} + R_{t,s} - U \times ((4 - P_{t1,n,g} - P_{t2,n,g}) - A _{t2, t1} - B_{t,s}) \\ \forall s \in S_{t}, \forall t1 \in T, \forall t2 \in (T - \{t1\}), \forall g \in GPU_{n}, \forall n \in N
\end{split}
\end{align}

This MILP formulation is complex because it spans and unifies three different system decisions in our setting. 
Our Joint Optimizer constructs all the constraints automatically for a given instance and provides them to Gurobi~\cite{gurobi}. We use the PuLP interface for Gurobi to keep all variables within a single Python process space.

\subsection{\textbf{Simulation-based Comparisons}}\label{sec:sched_eval}

We now evaluate our MILP-solver approach. 
We begin by discussing baselines from current practice and heuristics in prior art. 
Then, we run evaluations on simulated workloads and find that the MILP-solver outperforms the other approaches by a significant margin. 

\subsubsection{\textbf{Baselines}}\label{sec:milp_baselines}
As the case study in Section~\ref{sec:intro} highlighted, large-model users must currently tackle the \spase~problem manually.
So we can define the initial baseline based on current best practices. A common heuristic is to just maximize each task's allocation. 
Each task is given all GPUs in a node; then the best parallelism for that particular setting is applied. The models are run one after another. 
This optimizes \textit{local} efficiency and maximizes available GPU memory for each task. This heuristic becomes a suboptimal degenerate case of the apportioning and scheduling parts of the \spase~problem.
We call this baseline ``Max-Heuristic'', and anecdotally we find this is common in current practice. 

The opposite extreme would be to minimize the number of GPUs assigned to each task to maximize task-parallelism~\cite{mpms2021}. We call this baseline ``Min-Heuristic.'' While it runs many models in parallel, this approach suffers a lot of DRAM spilling for large models.

Finally, we devise a strong algorithmic heuristic that incorporates our runtime estimates to produce non-trivial solutions.
It extends an idea from Optimus, a DL resource scheduler in prior art that proposes a greedy resource allocator that uses an ``oracle'' to provide runtime estimates~\cite{peng2018optimus,qiao2021pollux}. 
Optimus iteratively assigns GPUs to whichever model that will see the greatest immediate benefit. 
\red{The original Optimus implementation used a throughput-prediction oracle for PS-style data parallelism, but subsequent works~\cite{qiao2021pollux} have made it standard to provide an alternate oracle to adapt Optimus for different parallelisms.}
Our Trial Runner statistics serve as our oracle, \red{thus allowing us to manually configure optimal parallelism selections for Optimus' benefit}. 
\red{Since this is not part of the base offerings of Optimus, we denote this strengthened modification of Optimus as Optimus*.
Optimus* serves as a strong baseline \spase~solver, tackling problems of resource allocation and model selection natively, and parallelism selection through our augmentation. \system's main advantage over this baseline
is its use of \textit{joint} optimization}. For our simulation study, we call this baseline algorithm Optimus*-Greedy. Algorithm~\ref{alg:optimus_greedy} presents its pseudocode, reusing variables from Table~\ref{table:milp_notation}.

\begin{algorithm}[h]
\caption{:~\textsc{\textbf{Optimus*-Greedy}}(Tasks $T$, GPUs $G$)}
\label{alg:optimus_greedy}
\begin{algorithmic}[1]
\STATE $L=[1 | t \in T]$
\WHILE{sum(L) < G }
	\STATE $CR = [R_{t,s} | t, l \in (T, L), s \in S_{t} \text{ where } G_{s,t} == l]$
	\STATE $PR = [R_{t,s} | t, l \in (T, L), s \in S_{t} \text{ where } G_{s,t} == l + 1]$
	\STATE $GAIN = [c - p | c, p \in (CR, PR)]$
	\STATE $L[ArgMax(GAIN)]++$
\ENDWHILE
\RETURN $L$
\end{algorithmic}
\end{algorithm}

The Optimus*-Greedy algorithm yields resource allocations per task. 
We transform that into a \spase~solution by selecting the best parallelism for each task's allocation post-hoc. 
In the multi-node case, we run this algorithm one node at a time. 
Like many iterative greedy algorithms, this approach relies on consistent scaling behaviors. 
It has only a local greedy view, rather creating a one-shot global resource distribution.

Apart from the above three approaches to cover standard practice and prior art extensions, we also include a simple randomization-based baseline. In summary, we compare with 4 approaches: 

\begin{enumerate}
\item \textbf{Max-heuristic:} All GPUs within a node are given to one task at a time.
\item \textbf{Min-heuristic:} A single-GPU technique (spilling) is given to each task to maximize task parallelism. If additional GPUs are available, they are divided evenly.
\item \textbf{Optimus*-Greedy:} A greedy algorithm inspired by the one used in the Optimus~\cite{peng2018optimus} resource scheduling paper.
\item \textbf{Randomized:} Parallelisms and allocations are randomly selected for every task, then tasks are randomly scheduled.
\end{enumerate}

For each of the above approaches, we use our Profiler results to select the best possible parallelism+allocation for each model. For instance, if a baseline determines that Model A should receive 8 GPUs, we refer to the Profiler to determine which parallelism gives Model A the best runtime at 8 GPUs.
This same best-check procedure is used to determine the gain values for Optimus*-Greedy.

Since our MILP is complex, Gurobi is unlikely to converge to an optimal solution in a practical timeframe. 
Thus, we set a reasonable timeout --- from our trials~\cite{techreport}, we set it to 5mins --- for the solver to produce a solution. 
\red{We rely on Gurobi's industry-strength techniques to find a high-quality (though possibly suboptimal) solution even within the allotted time.
The Appendix of our tech report~\cite{techreport} shows the diminishing returns of having a larger timeout.}
We leave it to future work to adapt the timeout for the given workload.

\subsubsection{\textbf{Simulation Workloads}}\label{sec:milp_sim}

\begin{figure}
\includegraphics[width=0.95\columnwidth]{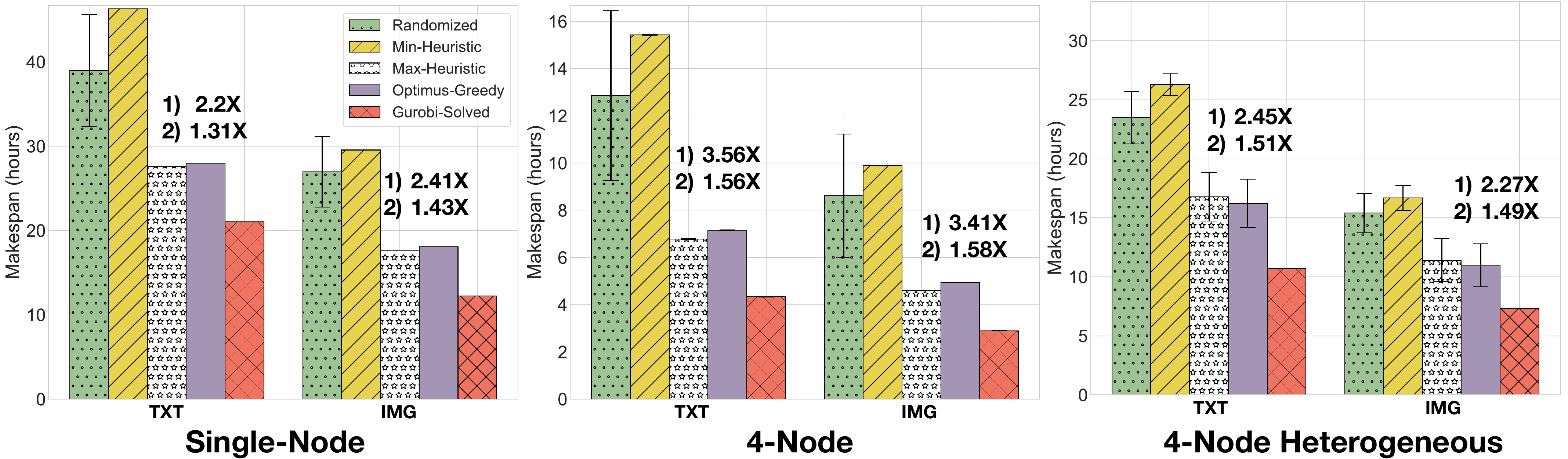}
\vspace{-3mm}
\caption{Simulation results comparing our MILP to two key baselines. For each group, we list \system's speedup versus (1) the weakest and (2) the second-best performer.}
\label{fig:simulations}
\vspace{-4mm}
\end{figure}

We simulate 2 benchmark workloads, described in Table~\ref{table:model_selection}. 
Runtime estimates for all models and configs are produced by the Trial Runner beforehand. 
We simulate 3 hardware settings: an 8-GPU single node, 32-GPUs over 4-nodes, and 4 heterogeneous nodes with GPU counts of 2, 2, 4, and 8 (16 GPUs in total). 
To adapt the baselines for the heterogeneous setting, we distribute models across nodes randomly, weighting each node's probability by its GPU count. 
Figure~\ref{fig:simulations} presents the simulation results. 
All approaches are run 3 times and averaged, with 90\% confidence intervals displayed; but only the randomized algorithm shows significant non-determinism on the homogeneous node settings.
In all cases, the MILP-solver approach yields significantly better solutions than the baselines. 
We achieve a makespan reduction of up to 59\% over the Min-Heuristic, 36\% over the Max-Heuristic, 54\% over Randomized, and 33\% over Optimus*-Greedy. 
In the heterogeneous setting, the improvements are slightly lower, ranging from 18\% to 42\%. 
We attribute this to the small 2-GPU nodes, which provide less flexibility for resource apportioning or parallelism selection, thus reducing the candidate solution space. 
Overall, \system's Gurobi-solved approach consistently outperforms the alternatives. 
The MILP-solved approach has the highest overhead; a 5min timeout versus $<10$ seconds for the baselines. 
But the overhead is negligible given the typical scale of the makespans. 

\subsection{\textbf{Introspection}}\label{sec:dynamic_rescaling}

In general, one-shot up-front scheduling is suboptimal.
Workloads can evolve over time, either due to online changes (e.g., an AutoML heuristic killing or adding models to train) or ongoing execution (task runtime reduce as they are trained).
If the optimizer can be rerun partway through execution, it might produce a different, more performant, solution for the remainder of the workload. 
To achieve this, we propose the use of \textit{introspection}~\cite{xiao2018gandiva}.

\begin{figure}
\includegraphics[width=0.85\columnwidth]{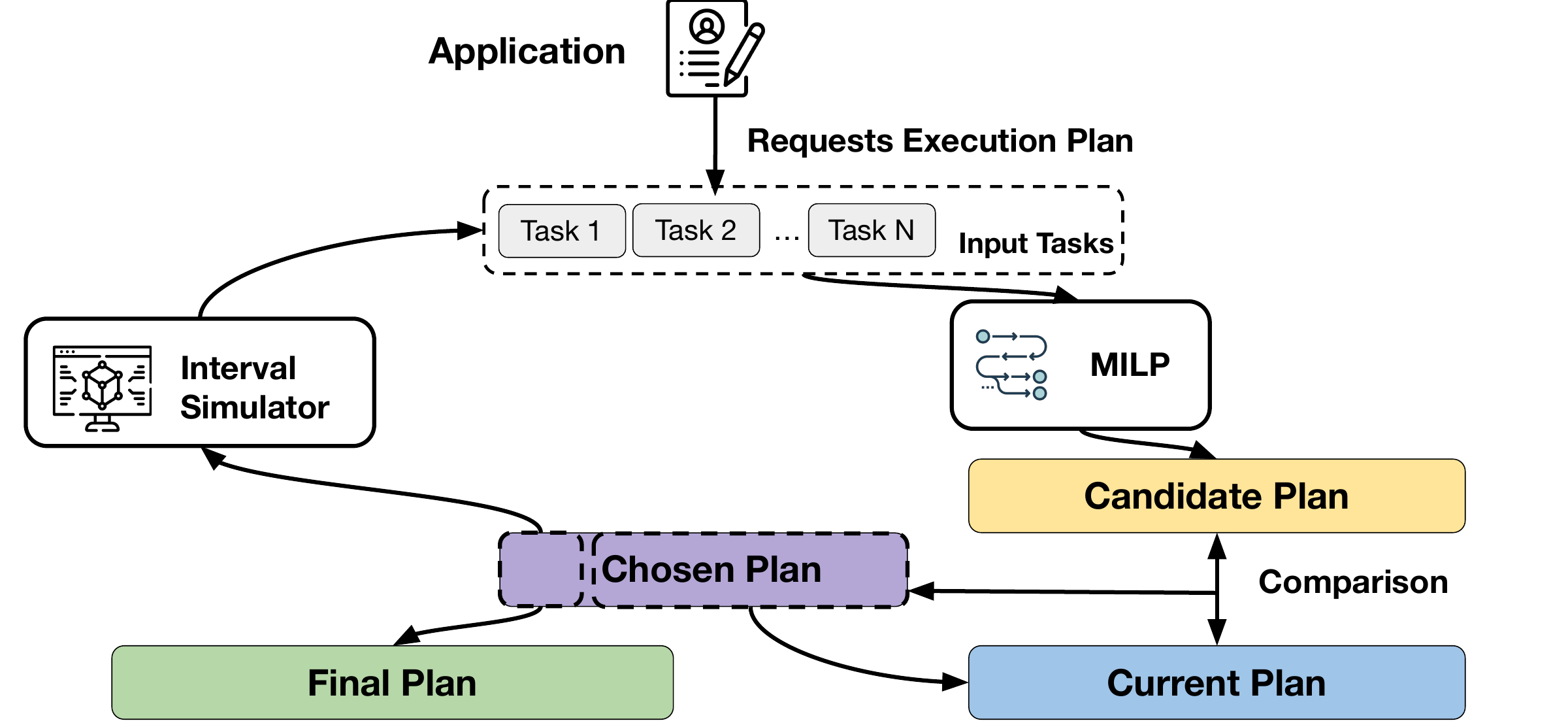}
\caption{Depiction of the introspective feedback loop.}
\label{fig:dynamic_rescaler}
\end{figure}

A key feature in some state-of-the-art DL schedulers~\cite{xiao2018gandiva}, introspection proposes that a scheduler should ``learn'' as it executes. 
There are two ways in which a schedule might be altered or adapted via introspection. First is pre-emption. 
Rather than blocking a GPU for a full job lifecycle, jobs can be swapped to different GPUs or paused temporarily. 
This enables fine-grained schedule construction and increased optimization flexibility. 
Second is dynamic rescaling. The initial up-front training plans could be adjusted (e.g., 6 GPUs down to 2) partway through a schedule. In \spase, this can also involve changing the parallelism.


We now describe how we implement introspection in \system. 
Figure~\ref{fig:dynamic_rescaler} illustrates our design.
We treat our \spase~MILP solver as a blackbox sub-system. 
At periodic intervals (e.g., every 1000 seconds), we re-evaluate the underlying workload. 
The partial training over the previous interval may have modified the set of models. 
We \textit{rerun} the solver on the interval boundaries so that it can introspectively adjust its original solution. 
By treating each interval-defined segment of training as effectively independent, we preserve gang scheduling semantics \textit{within} each segment, while allowing for graceful exits and relaunches across intervals. 
Such sequences of independent segments are possible due to the iterative nature of SGD, as well as the ease of checkpointing models during training~\cite{qiao2021pollux}. 
\red{Global batch size consistency is respected by adjusting per-device batch sizes to account for new allocations. 
Since we focus on model selection with the fidelity desideratum, we cannot modify the user-configured batch size transparently.}
Due to space constraints, we provide the full pseudocode of our approach in the Appendix of our tech report~\cite{techreport}.

\begin{figure}
\includegraphics[width=0.9\columnwidth]{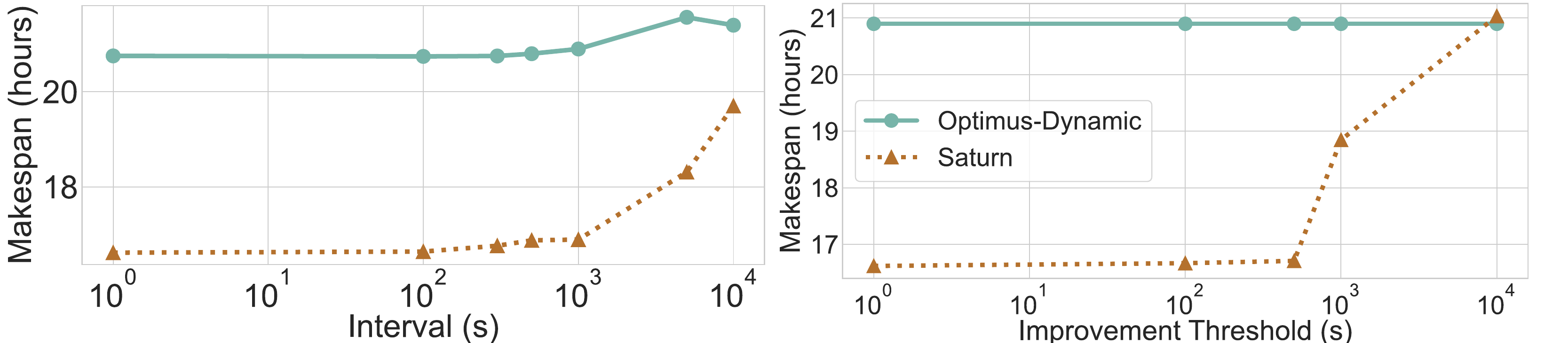}
\vspace{-3mm}
\caption{Sensitivity plots for \system~and Optimus*-Dynamic for interval and threshold knobs. We fix the interval to 1000s for the first analysis and the threshold to 500s for the second.}
\label{fig:scheduler_knobs}
\vspace{-5mm}
\end{figure}

To demonstrate the impact of \system's introspection, we compare with a new dynamic baseline, ``Optimus*-Dynamic'', by swapping the MILP-solver for the Optimus*-Greedy algorithm. 
Figure~\ref{fig:scheduler_knobs} shows the impact of the interval length and the improvement threshold knob. 
Since each round produces a holistically optimized solution, \system's performance improves monotonically (not accounting for pre-emption costs) as knobs become more fine-grained. 
Lower interval/threshold levels naturally subsume higher levels in this scheme. 
In contrast, locally-optimizing algorithms such the Optimus*-Dynamic approach have non-monotonic behaviors. 

Introspection does not have to occur on interval completion; we can simulate the next-interval state based on the current solution.
Then, the solving process for the next introspection round can be overlapped with execution of the current round to hide the latency of introspection. 
This scheme provides speedups of 15\% to 20\% over our one-shot MILP, as shown in Section~\ref{sec:ablation}.
With introspection plus our MILP-solver, \system's Joint Optimizer is 1.5x-4.1x faster than the heuristics described in Section~\ref{sec:sched_eval}. 
Our introspection optimization significantly improves offline execution, but it also naturally supports online AutoML optimizations such as early-stopping~\cite{hyperband,asha2018} \red{or new job arrivals in a multi-tenant cluster} through workload reassessment. 
We do not explicitly optimize for AutoML heuristics in the current version of \system; but it is easy to extend it to exploit this optimization.


Our introspection optimization takes inspiration from prior art in DL cluster scheduling, e.g., Antman~\cite{xiao2020antman} and Gandiva~\cite{xiao2018gandiva} which demonstrated the value of pre-emption on minibatch boundaries, as well as Pollux and Optimus~\cite{qiao2021pollux,peng2018optimus}, which showed the value of dynamic rescaling. 
Our contribution is in unifying both of those optimization ideas to craft our introspection technique, which also incorporates change-of-parallelism across introspection rounds. 

%

%

\begin{table*}[t]
\centering
\caption{Model selection configurations of workloads.}
\label{table:model_selection}
\scalebox{1.0}{
\begin{tabular}{p{1.5cm} c c c c p{1.5cm} c}
\toprule
\multirow{2}{*}{Workload} & \multicolumn{5}{c}{Model Selection Configuration} & \multirow{2}{*}{\# Models}\\
\cmidrule{2-6}
 &  Model Arch. (params) & Dataset & Batch Size & Learning Rate & Epochs & \\
\midrule
TXT & {GPT-2 (1.5B), GPT-J (6B) } & WikiText-2 & \{16, 32\} & \{1e-5, 1e-4, 3e-3\} & 10 & 12\\
IMG & {ViT-G (1.8B), ResNet (200M)} & ImageNet & \{64, 128\} & \{1e-5, 1e-4, 3e-3\} & 10 & 12\\
\bottomrule
\end{tabular}
}
\end{table*}

\section{Experimental Evaluation}\label{sec:experiments}
We now run an extensive empirical evaluation. We aim to answer two questions: (1) What performance benefits does~\system~provide compared to current practice? (2) How much do each of~\system's optimizations contribute to the overall speedups?

\noindent \textbf{Workloads, Datasets, and Model Configurations:} 
We run 2 model selection workloads with benchmark DL tasks. 
Table~\ref{table:model_selection} lists the model selection configurations for both workloads. 
The first (TXT) is a text workload with LLMs. It uses the popular \textit{WikiText-2}~\cite{wikitext-2} dataset. 
WikiText-2, which is drawn from Wikipedia, has previously been used as a benchmark on landmark LLMs such as GPT-2~\cite{radford2019language}.
TXT uses two GPT models: GPT-2 (1.5B parameters), introduced in 2019, and GPT-J (6B parameters), introduced in 2021. Both are still considered state-of-the-art for application-specific finetuning purposes. 
The second (IMG) is image classification comparing a large ResNet (200M parameters) and a large-scale Vision Transformer (1.8B parameters). 
It uses the computer vision benchmark dataset \textit{ImageNet}~\cite{imagenet} (14M images and 1000 classes). 

\noindent \textbf{Software Setup:} 
All models are implemented and trained with PyTorch 2.0. 
We register 4 parallelisms in \system. 
\begin{enumerate}
\item PyTorch Distributed Data Parallelism~\cite{torchddp2020}.
\item PyTorch Fully-Sharded Data Parallelism~\cite{torchddp2020}.
\item GPipe, adapted from an open-source implementation~\cite{torchgpipe2020}.
\item Model spilling, provided by the FairScale library~\cite{FairScale2021}.
\end{enumerate}

We use Gurobi 10.0 for our \spase~MILP-solver; the introspection threshold and interval parameters are set to 500s and 1000s, respectively. For the underlying job orchestration, we use Ray v2.2.0. Datasets are copied across nodes upfront. 

\noindent \textbf{Hardware Setup:} 
We configure 3 hardware settings: (1) 8-GPU single-node, (2) 16-GPU 2-nodes, and (3) heterogeneous 2-nodes, where one node has 8 GPUs and the other has 4 (12 GPUs in total).
All settings use AWS p4d instances.

\noindent \textbf{Baselines:} 
No prior end-to-end system can solve the \spase~problem; prior art either does not support large models or else fails model selection constraints, as Table~\ref{tb:prior_art} showed.
So, we compare \system~with 4 baselines using the approaches in Section~\ref{sec:sched_eval}.

\begin{enumerate}
\item[(1)] \textbf{Current Practice:} A heuristic without any task parallelism within nodes. It allocates 8 GPUs per task. 
Parallelism selection is set by a human to ``optimal'' choices for an 8-GPU allocation, (typically FSDP). 
This is perhaps most representative of current practice by end users of DL. 

\item[(2)] \textbf{Random:} A randomizer tool selects parallelism and apportioning and then applies a random scheduler. This represents a system-agnostic user.

\item[(3\&4)] Two modified versions of Optimus*-Greedy (Alg. 1) combined with a randomized scheduler (see Section~\ref{sec:optimizations}). 
We name these baselines \textbf{Optimus*-Dynamic} and \textbf{Optimus*-Static}. 
These are the strongest baselines for large-model model-selection we could assemble from prior art. 
\end{enumerate}

The above approaches cover both current practices and reasonable strong heuristics for our problem setting. 
We note that the two Optimus*-based baselines use our Trial Runner as an oracle for their runtime estimates \red{and parallelism selection decisions} (the original Optimus paper only had runtime models for Parameter Server-style data parallelism~\cite{peng2018optimus}). 
This highlights the novelty of our problem setting --- the strongest baseline from prior art needs to reuse a module of our system.


\subsection{End-to-End Results}\label{sec:experiments_end_to_end}

\begin{figure*}[t]
\includegraphics[width=\textwidth]{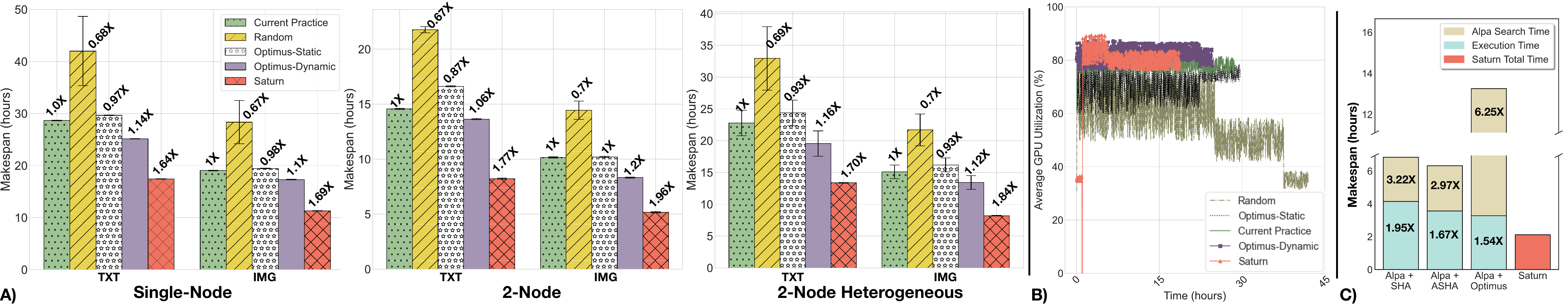}
\vspace{-6mm}
\caption{(A) End-to-end runtimes. Speedups versus current practice are also noted. Results are averaged over three trials, with the 90\% confidence interval displayed.
(B) Average GPU utilization over time at a 100s sampling rate on the single-node TXT job. (C) End-to-end runtimes of \system~versus compositions of tools on a reduced version of the TXT job on 2 nodes.}
\label{fig:e2e}
\vspace{-2mm}
\end{figure*}

\noindent \textbf{Model Selection Runtimes:} 
We first compare the end-to-end runtimes versus the 4 baselines. 
The Trial Runner search overheads are \textit{included} in \system's runtime.
Figure~\ref{fig:e2e}(A) presents the results. 

\system~achieves significant speedups versus all baselines.
Against Current Practice, we see makespan reductions of 39-40\% on a single-node, 43-48\% on 2 homogeneous nodes, and 41-45\% on 2 heterogeneous nodes.
Against the strongest baseline (Optimus*-Dynamic), we see makespan reductions of 30-34\%, 38-40\%, and 32-39\% on the three hardware configurations respectively.
Since the same UPP implementations are used \textit{in all cases}, the speedups are achieved purely via the better parallelism selections, resource allocations, and schedule constructions. 
All compared approaches (including \system) use logically equivalent SGD and offer the same accuracy. 

Figure~\ref{fig:e2e}(B) plots GPU utilization.
\system~achieves good utilization throughout, except an initial low-utilization period for the Trial Runner's search and MILP solving period. 
GPU utilization alone can be misleading; tools such as nvidia-smi can artificially inflate utilization~\cite{nvidiasmi}. So, these results should not be taken as a measure of training performance in isolation.

Overall,~\system~reduces model selection runtimes substantially for all workloads in all evaluated settings. 
It also offers more qualitative benefits to end users of DL because they are freed from manually selecting parallelisms, deciding on resource allocations, or tuning system parameters.

\noindent \textbf{Intuition on Efficiency Gains.} 
\system's performance improvements arise due to its \textit{holistic} optimization approach. 
To the best of our knowledge, this is the first work that characterizes the parallelism performance crossovers and incorporates them into a joint optimizer.
Our empirical profiler and unified \spase~formulation enable us to optimize in a parallelism-agnostic fashion. 
The heuristic and algorithmic baselines make assumptions about scaling behaviors (e.g., consistency, linear scaling, etc.) that do not always hold up in large-model DL practice. 
To prove our point further, Table~\ref{table:selected_outputs} lists the parallelisms+allocations selected by \system~for a few models from the single-node workloads. We see a non-trivial mixture of decisions across the models trained.

\begin{table}[h]
\centering
\caption{Parallelisms and apportionments chosen by \system~for a few evaluated models.}
\label{table:selected_outputs}
\begin{tabular}{clclc}
\toprule
Model Config & Parallelism & Apportionment\\
\midrule
GPT-2 (Batch 16, 1e-5 LR) & Pipelining & 5 GPUs \\
GPT-2 (Batch 32, 1e-4 LR) & FSDP & 4 GPUs\\
GPT-J (Batch 16, 1e-5 LR) & FSDP & 8 GPUs\\
GPT-J (Batch 32, 1e-4 LR) & Pipelining & 3 GPUs \\
ResNet (Batch 64, 1e-4 LR) & DDP & 2 GPUs \\
ResNet (Batch 32, 1e-4 LR) & Spilling & 1 GPU\\
ViT-G (Batch 32, 1e-4 LR) & FSDP & 4 GPUs\\
ViT-G (Batch 16, 1e-4 LR) & FSDP & 6 GPUs\\
\bottomrule
\end{tabular}
\end{table}

\system's MILP-chosen \spase~solutions combine into a multi-model \spase~solution to minimize end-to-end runtimes.
Our unified data systems-style approach frees DL users to focus on their goals instead of tedious low-level decisions.

\subsection{Joint Optimization Evaluation}
To better understand the value of joint optimization for \spase, we evaluate \system~against different compositions of tools --- Alpa~\cite{alpa2022} + ASHA~\cite{asha2018};  Alpa + Optimus*~\cite{peng2018optimus}; Alpa + SHA~\cite{hyperband} --- each used together but unaware of each other. A/SHA \& Optimus are designed for multi-model training and GPU allocation; Alpa tackles parallelism selection. In combination, they can be used to solve the dimensions of the \spase~problem, but in a separated fashion. We elaborate on these tools in Section \ref{sec:related_work}.

To mimic A/SHA's early-stopping behaviors, we run \system~and Optimus* one epoch at a time. We take the early stops produced by A/SHA and apply them to \system~and Optimus*'s workloads on epoch boundaries.
A/SHA is configured to use 3 rungs, with allocations of 1, 3, and 6 epochs respectively, so completed jobs will have run for 10 epochs.
The decay factor is set to 2, so half of the jobs survive each rung. 
Since A/SHA was built for settings with substantially more accelerators than models, we use a smaller version of the TXT workload with 8 jobs (eliminating the 3e-3 learning rate option) on 2 nodes.

We report the results in Figure~\ref{fig:e2e}(C). We find that \system~outperforms Alpa + ASHA by nearly 3X. Even if we remove Alpa's search times (e.g., if the searches were run once up-front) and directly compare \spase~solution quality, \system~still outperforms the composite baseline by 1.67X. Against Alpa + Optimus*, the speedups are 6.25X (resp. 1.54X) when including (resp. excluding) Alpa's compilation times. The Optimus* runtime that includes the compilation times is so high because it needs to construct its throughput oracle~\cite{qiao2021pollux} up front by running the compiler for every possible allocation for every model. \system's significant speedups against all 3 baselines support our view that the \spase~problem is a novel space where joint optimization has a significant role to play, rather than a simple composition of existing problem spaces.

\subsection{Drilldown Analyses}\label{sec:ablation}

\subsubsection{\textbf{Ablation Study}}  We separate our optimizations into 4 layers: scheduling, resource allocation, parallelism selection, and introspection. 
We apply these one-by-one as follows. 
First, a version without any of our optimizations. FSDP is used with checkpointing and offloading (i.e., a non-expert config), resource allocations are set manually to 4 GPUs per task, and a random scheduler is used.
Second, we use our makespan-optimized scheduler. 
Third, we reintroduce resource apportioning to the MILP.
Fourth, we allow for automatic parallelism selection and knob tuning.
Finally, we overlay introspection. This completes~\system.
We use the single-node TXT workload in our study.
Table~\ref{tb:ablation} notes the marginal speedups.

\begin{table}[h]
\centering
\vspace{-3mm}
\caption{Ablation study.}
\label{tb:ablation}
\vspace{-4mm}
\begin{tabular}{clclc}
\toprule
Optimizations & Abs. Speedup & Extra Speedup\\
\midrule
Unoptimized & 1.0X & 1.0X\\
+ MILP Scheduler & 1.1X & 1.1X\\
+ Resource Allocation in MILP & 1.33X & 1.2X\\
+ Auto. Parallelism Selection & 1.95X & 1.47X \\
+ Introspection & 2.27X & 1.16X \\
\bottomrule
\end{tabular}
\vspace{-1mm}
\end{table}

The scheduler-only MILP provides better packing for some initial makespan improvements. 
Adding in resource apportioning lets the solver reshape task runtimes and demands to produce more speedups. 
Automatic parallelism selection creates even more flexibility and adds in knob-tuning to improve parallelism performance. Introspection enables the solver to reassess its solution and adapt to shifts in the workload to cap off \system's speedups.

\begin{figure}
\includegraphics[width=\columnwidth]{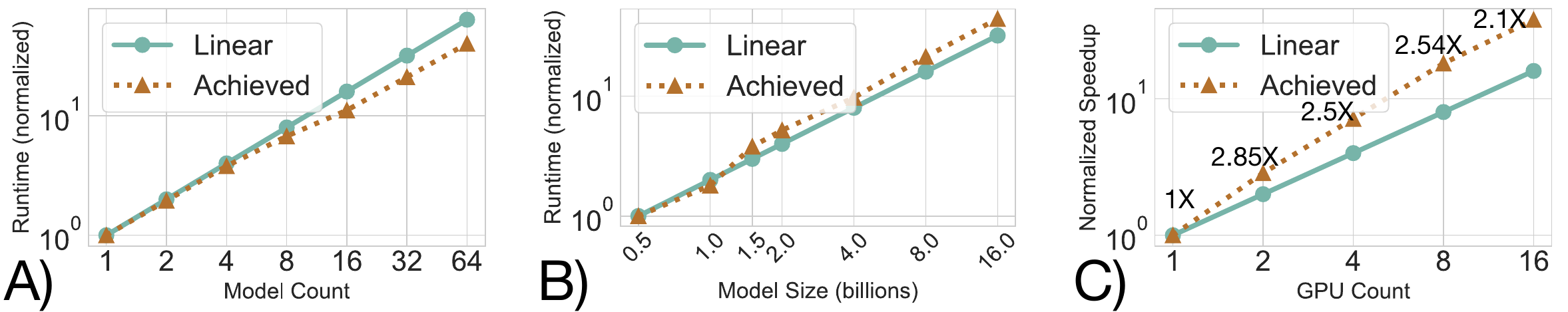}
\vspace{-6mm}
\caption{\system~sensitivity plots on the TXT workload versus (A) workload size, (B) model size, and (C) node size. Charts are in log-log scales, normalized to the initial setting. (C) labels each point with the marginal speedup.}
\label{fig:opt}
\vspace{-5mm}
\end{figure}

\subsubsection{\textbf{Sensitivity Analyses}} We test \system's sensitivity to the size of: (1) workloads, (2) models, and (3) clusters. 

For workload size scaling, we run TXT on a single 8-GPU node with the GPT-2 model, set batch size to 16, and vary the number of learning rates explored. Figure~\ref{fig:opt}(A) presents the results.
\system~scales slightly superlinearly as larger workloads enable broader scope for optimization.
This suggests strong performance on large-scale model selection workloads.

Next, we vary model size. We run TXT on a single 8-GPU node with batch size set to 16 and learning rate set to 1e-5.
All models are versions of GPT-2. We vary model size by stacking encoder blocks, akin to what GPT-3 does~\cite{gpt2020}. Figure~\ref{fig:opt}(B) presents the results.
\system~achieves mostly linear scaling, but with slight slowdowns on the largest model sizes. 
This is because the largest models force the \spase~solution to use the only viable configuration (8-GPU FSDP with checkpointing and offloading) for every model. 

Finally, we vary the number of GPUs visible to~\system. 
We use TXT for this experiment. Figure~\ref{fig:opt}(C) presents the results.
\system~achieves superlinear speedups for 2 reasons.
First, the single-GPU case necessitates DRAM spilling, while larger GPU counts reduce the spilling required and open up more parallelism options.
Second, higher GPU counts broaden the solution space for the MILP, enabling higher flexibility.

\section{Related Work}\label{sec:related_work}

\system's focus on the unified \spase~problem is a first for large-model DL workloads. 
We now elaborate on prior art connected to each dimension of the \spase~problem. The Appendix of our tech report includes further discussion of the wider DL systems landscape.

\noindent \textbf{Parallelism Selectors and Hybridizers:} 
Paleo~\cite{qi2016paleo} focused on performance models for data parallelism and model parallelism. 
But the DL parallelism landscape has changed since then (2016), with numerous new approaches. 
While Paleo might be extended to newer parallelisms, our empirical Trial Runner approach is more easily extensible and highly general. 
Alpa, FlexFlow, and Unity~\cite{alpa2022,flexflow2018,unger2022unity} focus on generating bespoke parallelism strategies for model architectures through complex search procedures. 
They can produce efficient \textit{single-model plans}, but the cumulative search overheads can get high when applied repeatedly to multi-model training. They also do not consider multiple models being trained in model selection workloads. 
In addition, these tools must manually be redesigned for new approaches (e.g., spilling). 
These tools could potentially be viewed as parallelisms for \system's UPP abstraction.

\noindent \textbf{DL Model Selection Systems:}
\system~follows a line of work on systems for model selection, including Cerebro~\cite{kumar2021cerebro}, Hyperband~\cite{hyperband}, and ASHA~\cite{asha2018}.
\red{However, none of these prior works were explicitly designed for the large-model setting, where users must navigate multiple complex and varied parallelisms, as explained in Section~\ref{sec:intro}.}
Cerebro hybridizes task- and data-parallelism to train multiple DL models in parallel on sharded data. 
Hyperband reallocates training resources (e.g., number of epochs) across tasks based on convergence behaviors. 
SHA implements a rung-based promotion plan to kill off less-promising job instances and prioritize the execution of higher-value ones.
ASHA extends this to execute promotions asynchronously. 
\red{ModelKeeper~\cite{modelkeeper2023} suggests warm-starting across similar model configurations. 
All these techniques exist at a higher-level of abstraction, e.g., data sharding, early-stopping, or warm-starting. 
Thus, they are orthogonal to \system~and could be combined with our work in future extensions.}

\noindent \textbf{DL Resource Schedulers:} 
Pollux and Optimus~\cite{qiao2021pollux,peng2018optimus,dl2} tackle apportionment and scheduling, two parts of \spase. But they do not explicitly support larger-than-GPU-memory models, where complex parallelisms alter performance tradeoffs in non-trivial ways, as our work shows. \red{In Optimus' case, we can take the core mechanisms and adapt them for large-model training, as we do in our Optimus* baselines. But such adaptations underperform native large-model tools such as \system.}
These tools also do not target model selection workloads and optimize for throughput, while makespan is better suited for our setting. They also alter model accuracy, violating our fidelity desideratum. A config submitted to Pollux (e.g., batch size X and learning rate Y) may yield different accuracies than the same X and Y without Pollux. 
Themis~\cite{mahajan2020themis} studies scheduling fairness for ML jobs from different users; their goal and setting are orthogonal to ours in that we focus on model selection jobs from the same user and optimize for makespan.

\section{Conclusions and Future Work}\label{sec:conclusions}

Finetuning of pre-trained large DL models is increasingly common. 
But navigating the complex space of model-parallel training is unintuitive for DL users even though it is needed to reduce runtimes and costs. 
The complex interplay of parallelism selection with model selection workloads, which requires resource apportioning and scheduling decisions, can also lead to high resource wastage if improperly handled. 
This work resolves these issues by formalizing the joint \spase~problem unifying large-model parallelism selection, resource apportionment, and scheduling and designing a new information system architecture we call \system~to tackle \spase. 
With user-friendly APIs, joint optimization, and a judicious mix of systems techniques, \system~reduces large-model DL model selection runtimes by 39-49\% over current practice, while freeing DL users from tedious systems-level decisions. 
Overall, \system~offers maximal functionality in a critical DL setting, while promoting architectural simplicity to ease real-world adoption.
\red{Future extensions could explore alternative algorithmic approaches to the \spase~problem, extend \system~for other scheduling objectives, and handle autoscaling clusters and dynamic job re-configurations.}

\begin{acks}
 This work was supported by a Meta Research Fellowship, an NSF CAREER grant, and gifts from VMWare. The content is solely the responsibility of the authors and does not necessarily represent the views of any of these organizations.
\end{acks}

\bibliographystyle{ACM-Reference-Format}
\bibliography{main}

\end{document}